\documentclass[graybox]{styles/svmult}

\usepackage{mathptmx}       
\usepackage{helvet}         
\usepackage{courier}        
\usepackage{type1cm}        
\usepackage{natbib}
\usepackage{amssymb}
\usepackage{amsmath}
\usepackage{makeidx}         
\usepackage{graphicx}        
\usepackage{multicol}        
\usepackage[bottom]{footmisc}

\makeindex             

\title*{A Unified 3D Mapping Framework using a 3D or 2D LiDAR}
\author{Weikun Zhen and Sebastian Scherer}
\institute{Weikun Zhen \at Carnegie Mellon University, \email{weikunz@andrew.cmu.edu}
\and Sebastian Scherer \at Carnegie Mellon University, \email{basti@andrew.cmu.edu}}
\begin{document}

\maketitle

\abstract{​Simultaneous Localization and Mapping (SLAM) has been considered as a solved problem thanks to the progress made in the past few years. However, the great majority of LiDAR-based SLAM algorithms are designed for a specific type of payload and therefore don't generalize across different platforms. In practice, this drawback causes the development, deployment and maintenance of an algorithm difficult. Consequently, our work focuses on improving the compatibility across different sensing payloads. Specifically, we extend the Cartographer SLAM library to handle different types of LiDAR including fixed or rotating, 2D or 3D LiDARs. By replacing the localization module of Cartographer and maintaining the sparse pose graph (SPG), the proposed framework can create high-quality 3D maps in real-time on different sensing payloads. Additionally, it brings the benefit of simplicity with only a few parameters need to be adjusted for each sensor type.
}

\section{Introduction}\vspace{-3mm}
\begin{figure}[t]
\centering
\includegraphics[height=1.3in]{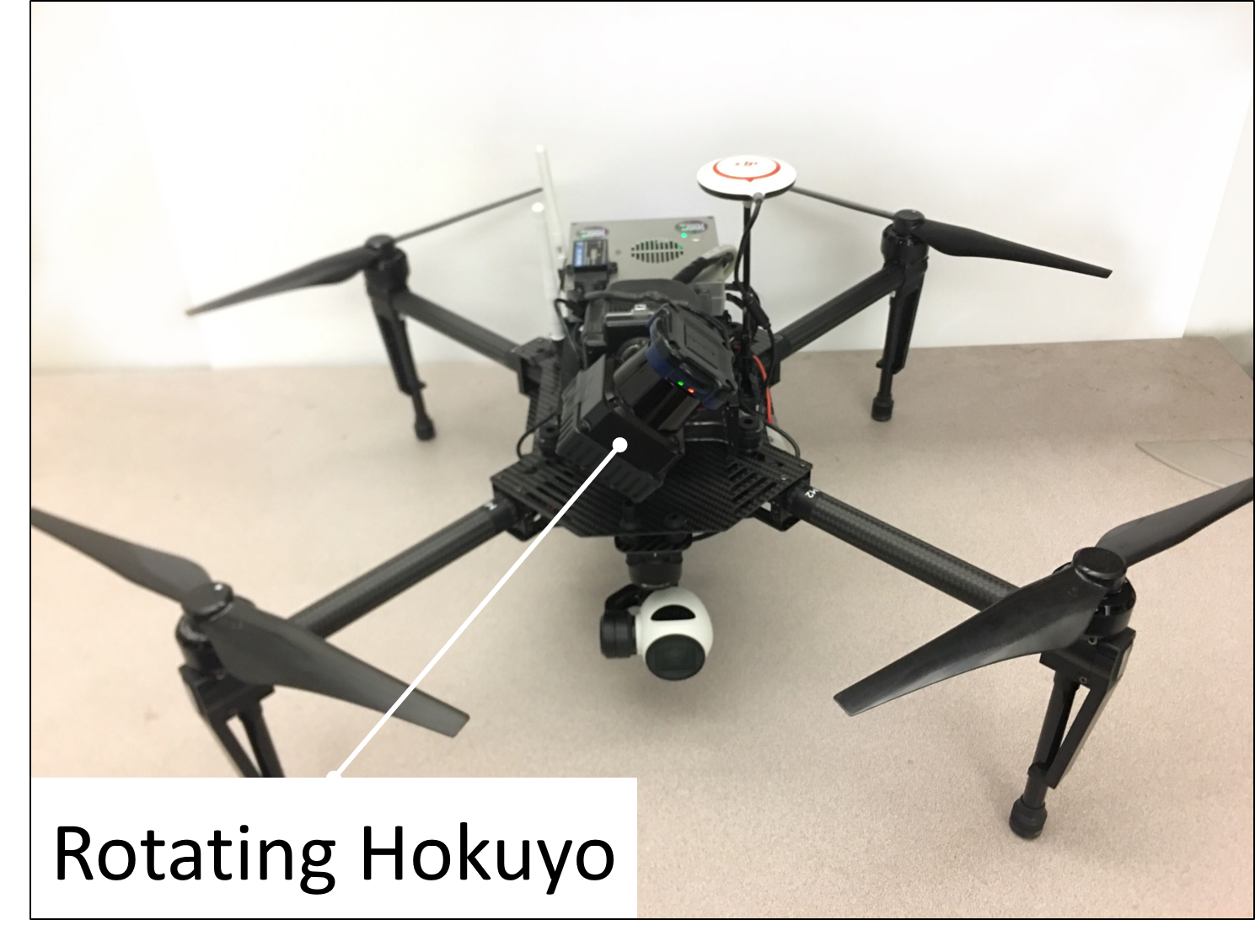}
\includegraphics[height=1.3in]{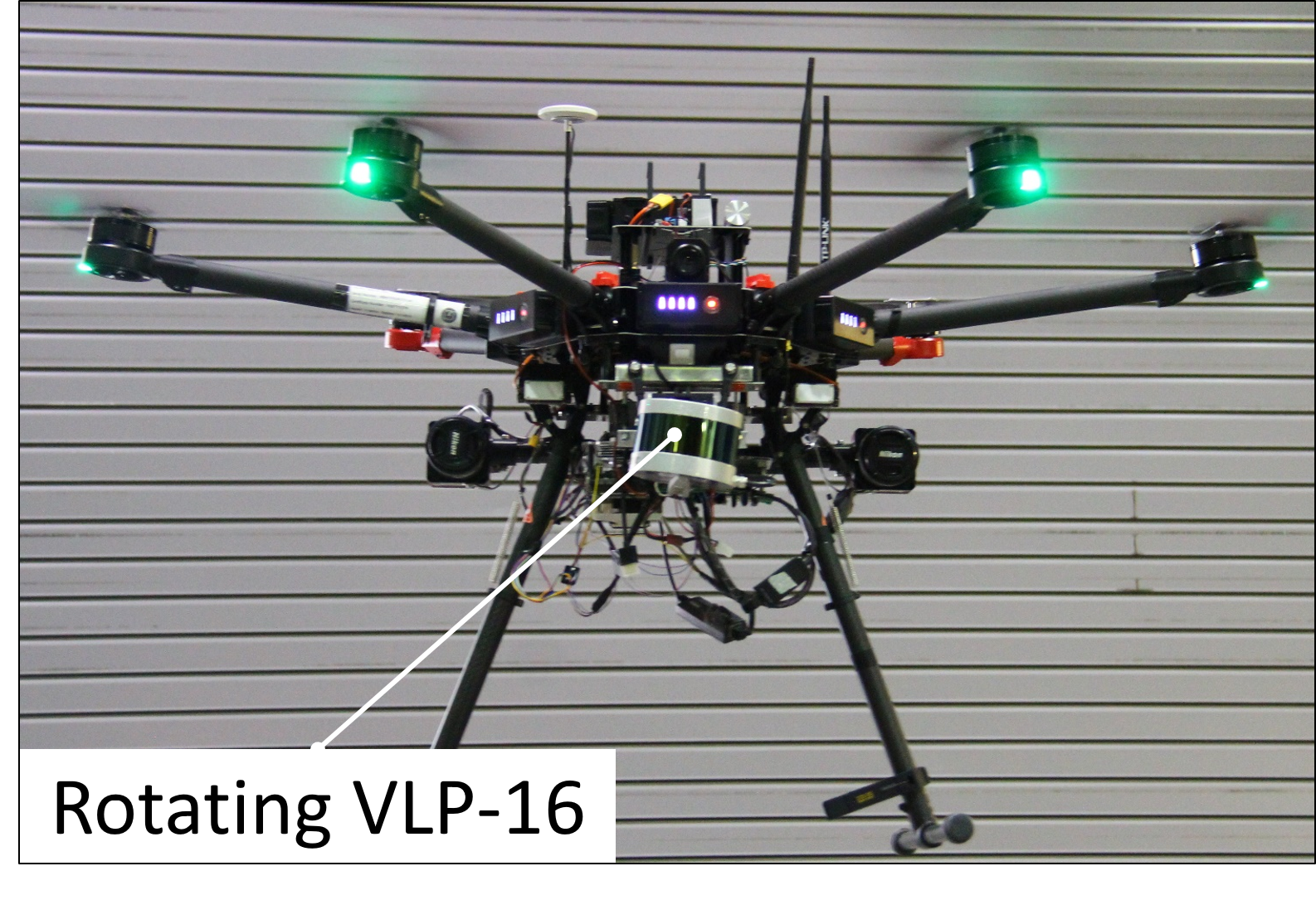}
\caption{\textit{Left}: A customized DJI M100 drone carrying a rotating Hokuyo laser scanner. \textit{Right}: A customized DJI M600 drone equipped with a rotating VLP-16.}\vspace{-5mm}
\label{fig:robot}
\end{figure}
​As one of the most fundamental problems of autonomous robots, LiDAR-based SLAM has been an active research area for many years. Recent advancements in LiDAR sensing and SLAM techniques have led to the fast growth of robot applications in many industrial fields such as autonomous inspection of civil engineering facilities. Usually, specialized systems are developed according to the particular requirements in different situations. For example, different Unmanned Aerial Vehicles (UAVs) (see Figure \ref{fig:robot}) are used to inspect tunnels in our project depending on the scale of the operating environment. The robots are equipped with different LiDARs and typically different SLAM algorithms would be required. However, maintaining multiple algorithms needs significant effort which is especially undesirable in the field. Consequently, in this work we propose a \emph{unified} mapping framework that handles multiple types of LiDARs including (1) a fixed 3D LiDAR, (2) a rotating 3D/2D LiDAR. 

The problem can be defined as follows: given a sequence of laser scans collected from a LiDAR sensor of any type, the algorithm will compute the motion of the sensor and build a 3D map in the meantime. As stated before, our work is based on the Cartographer SLAM \citep{carto} which contains a foreground localization component and a background SPG refinement component. Originally designed to work with stationary 3D LiDARs, it doesn't generalize to rotating 2D LiDARs since directly accumulating 2D scans using the IMU in the localization component will introduce distortion to the map. To accommodate this problem, we apply a different localization method that has two major advantages. First, every single scan is matched to the map to compute a more accurate pose than pure IMU-based methods. Second, the pose is computed regardless of the LiDAR types, allowing the framework to be generalizable across different platforms. With a unified framework, identical parameter-tuning strategies can be shared between systems which significantly simplifies the set-up procedure of multiple platforms during field tests. Additionally, we show that only a few parameters need to be adjusted when switching platforms such as local map resolution, number of accumulated scan and so on. More details will be discussed in the experiments.

The rest of this paper is structured as follows. Section \ref{sec:relatedwork} summarizes the related work on LiDAR-based SLAM problem. The proposed method is described in detail in Section \ref{sec:approach}. Experiments and results are presented in Section \ref{sec:experiments}. Finally, conclusions and insights are discussed in Section \ref{sec:conclusion}.

\section{Related Work} \vspace{-3mm}
\label{sec:relatedwork}

There has been a vast of research on LiDAR-based SLAM over past decades. Classic probabilistic approaches such as Kalman filters \citep{castellanos2012mobile} \citep{montemerlo2002fastslam} and particle filters \citep{dellaert1999monte} \citep{doucet2000rao} \citep{gmapping} infer the distribution of the robot state and the map based on measurements which are characterized by sensor noise models. \cite{thrun2002robotic} does a comprehensive review on the techniques. Those work establishes the theoretical fundamentals of the SLAM problem and has achieved great success in robustness, accuracy and efficiency. However, most of these approaches are limited to using fixed 2D LiDARs to solve the planar SLAM problem. Although in principle these algorithms are generalizable to 3D, the computational cost could become intractable as the dimension increases. 

In 3D situations, 2D LiDARs may be mounted on a rotating motor \citep{bosse2009continuous} \citep{zlot2014efficient} \citep{zhang2014loam} or a spring \citep{bosse2012zebedee} to build 3D maps. The additional degree of freedom significantly enlarges the sensor FOV, which, on the other hand, makes sequential scan matching impossible due to a lack of overlap. To account for this issue, a smooth continuous trajectory \citep{anderson2013towards} \citep{tong2013gaussian} may be used to represent robot motion instead of a set of pose nodes. However, the smooth motion assumption does not always hold true. 

More recently, as 3D ranging technology becomes widely used, methods to achieve real-time, large-scale and low-drift SLAM have emerged using accurate 3D LiDARs. \cite{martin2014two} developed a Differential Evolution-based scan matching algorithm that is shown to be of high accuracy in three dimensional spaces and contains a loop-closure algorithm which relies on surface features and numerical features to encode properties of laser scans. \cite{zhang2014loam} extract edge and planar features from laser scans and then adopt an ICP method \citep{chen1992object} for feature registration. An extension is presented in their later work \citep{zhang2015visual} where visual data is fused with range data to further reduce drifts. Although they do not compute the loop-closure, the generated map is of high accuracy even after travelling for several kilometers. \cite{carto} introduced the Cartographer SLAM where a local odometry relying on scan matching estimates the poses and meanwhile an SPG is updated and optimized regularly to refine pose estimates and generate consistent maps. Although existing methods vary in specific techniques, most share a similar pipeline, which estimates the pose using ICP or its variants as front-end while solves an SPG or trajectory optimization problem as the back-end. 

\section{Approach}
\label{sec:approach}\vspace{-3mm}
\subsection{Localization}\vspace{-3mm}
\begin{figure}[t]
\centering
\includegraphics[width=0.65\textwidth]{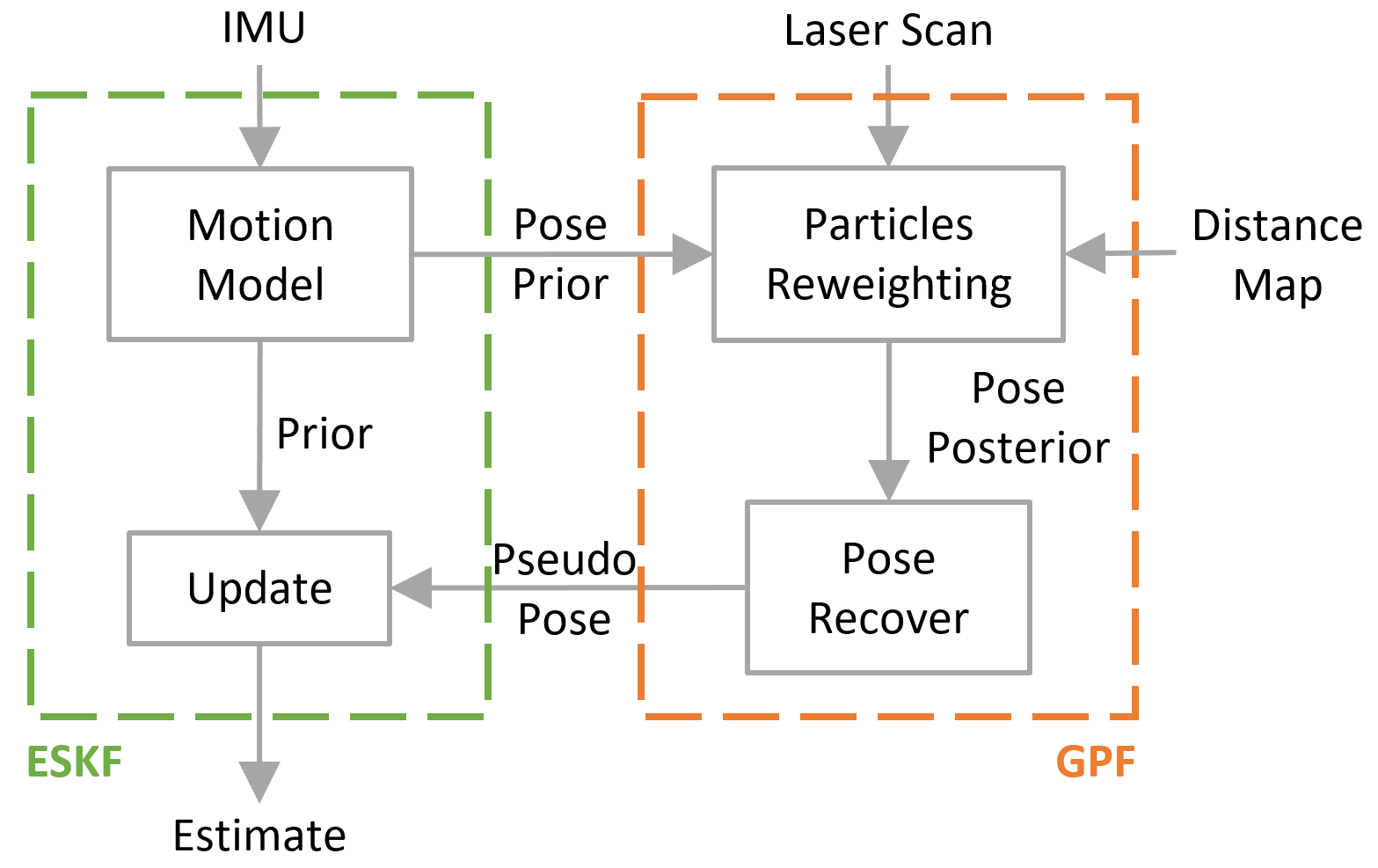}
\caption{Pipeline of the localization algorithm that takes the sensor readings and a distance map as input and outputs the pose estimate. }\vspace{-3mm}
\label{fig:local}       
\end{figure}
\begin{figure}[t]
\centering
\includegraphics[width=0.7\textwidth]{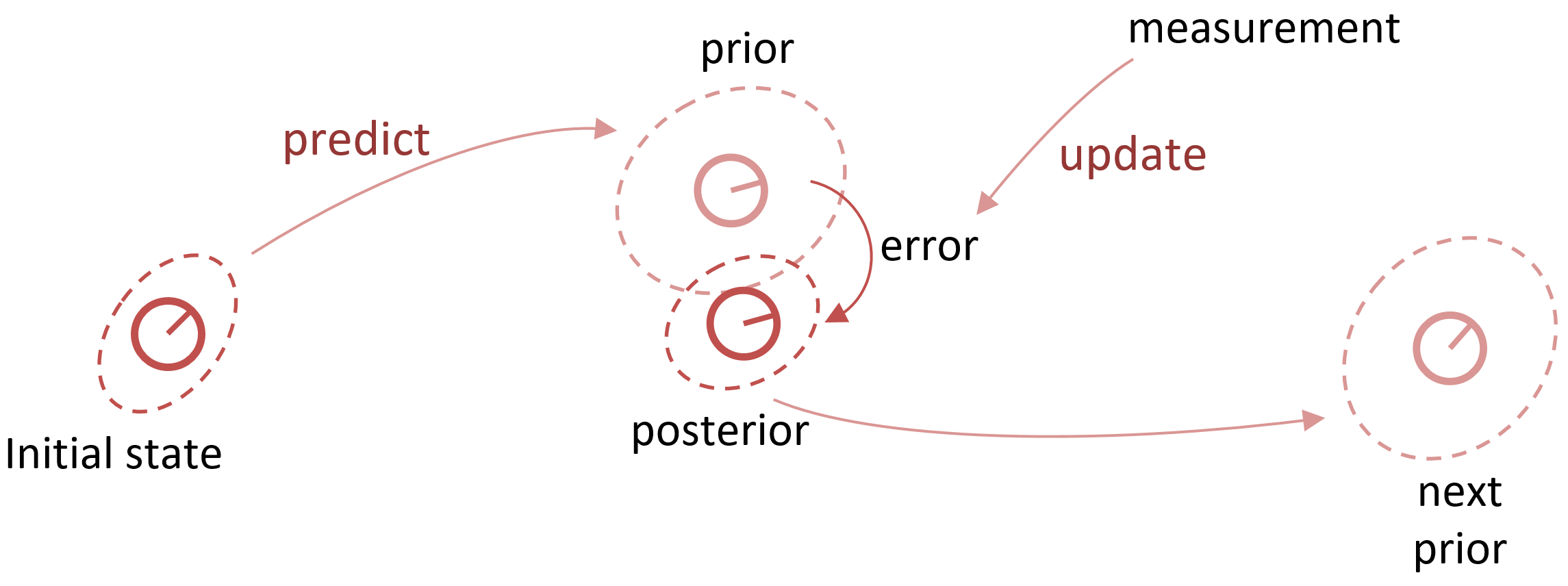}
\caption{An illustration of the ESKF in 2D. The circle denotes the robot with a bar indicating its orientation. The dashed ellipse represents the position uncertainty. Here the orientation uncertainty is omitted for simplicity.}\vspace{-3mm}
\label{fig:eskf}
\end{figure}
\begin{figure}[t]
\centering
\includegraphics[width=0.75\textwidth]{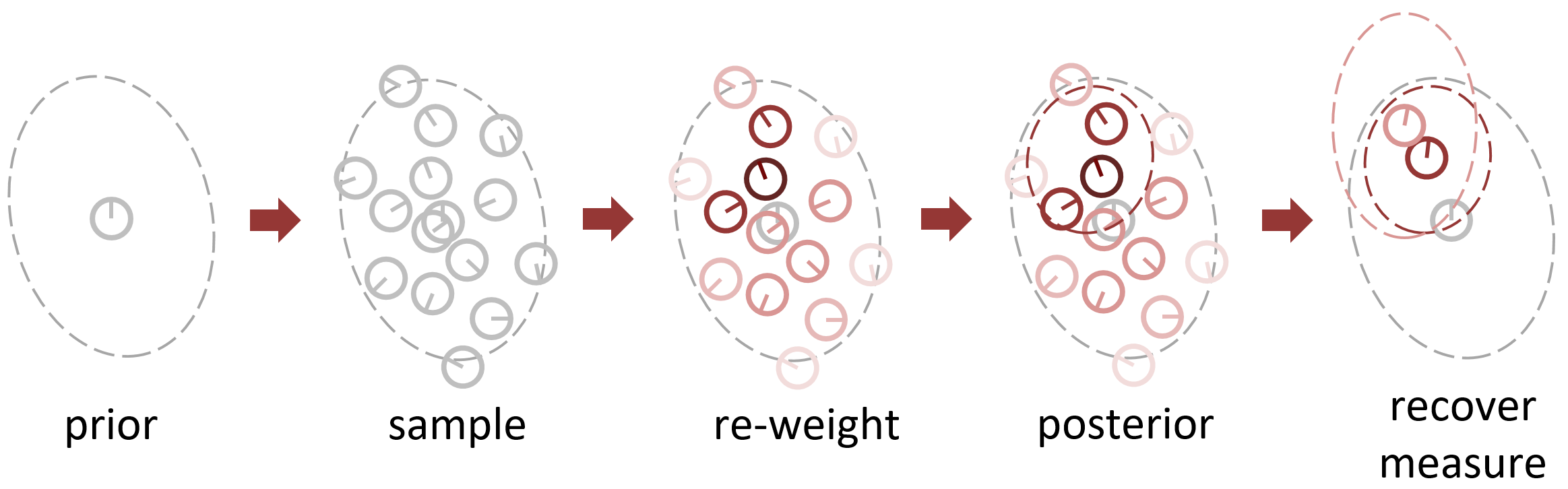}
\caption{An illustration of the GPF in 2D. Circles and ellipses share the same meaning as in Figure \ref{fig:eskf}. Differently, the darker color means a higher weight, namely higher probability of a hypothesis to be true.}\vspace{-3mm}
\label{fig:gpf}
\end{figure}

The localization module (shown in Figure \ref{fig:local}) combines an Error State Kalman Filter (ESKF) with a Gaussian Particle Filter (GPF) to estimate robot states inside a prior map. The GPF, originally proposed by \cite{bry2012state}, converts raw laser scans to a pose measurement, which frees the ESKF from handling 2D or 3D range data directly. This is a key factor that ensures compatibility. More specifically, the ESKF (illustrated in Figure \ref{fig:eskf}) numerically integrates IMU measurements to predict robot states and uses a pseudo pose measurement to update the prediction. In the GPF illustrated in Figure \ref{fig:gpf}, a set of particles (pose hypotheses) are sampled according to the prediction, then weighted, and finally averaged to find the posterior belief. By subtracting the prediction from the posterior belief, the pseudo pose measurement is recovered and used to update the ESKF. Finally, we refer the readers to our previous work \citep{eskf} for more details.


\subsection{Submaps and Distance Maps}\vspace{-3mm}
\begin{figure}[t]
\centering
\includegraphics[width=0.7\textwidth]{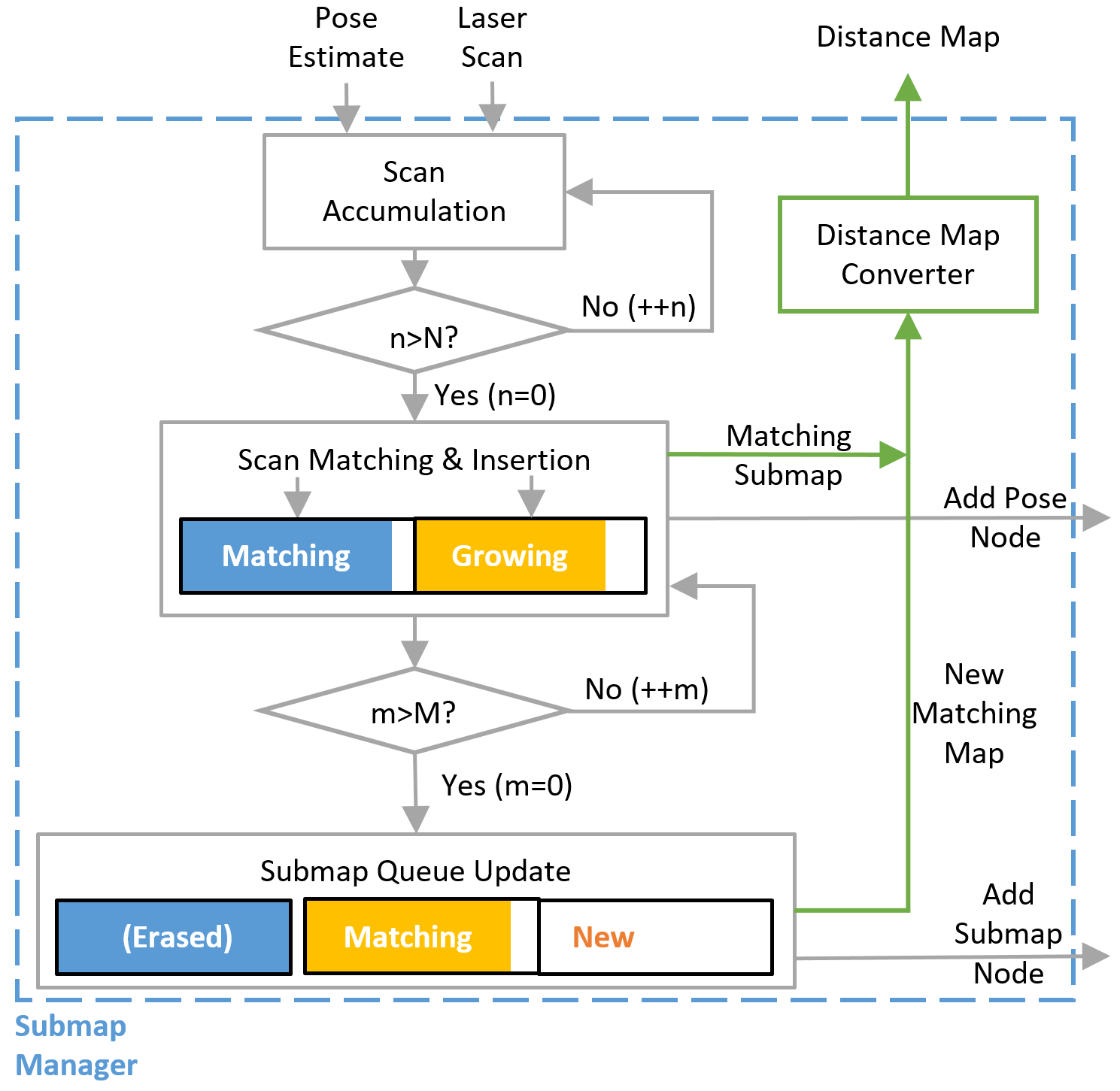}
\caption{The pipeline modified based on Cartographer to manage submaps and distance maps. Modifications are highlighted in green. 
}\vspace{-3mm}
\label{fig:graph}
\end{figure} 

A local occupancy grid map is defined as a submap by Cartographer. Since a different localization method is used, we need to adjust the submap management scheme so that the submaps can be accessed by the localization module. As shown in Figure \ref{fig:graph}, there exist two stages of scan accumulation. On the first stage, $N$ scans are accumulated to form a 3D scan, then matched and inserted into the active submaps. The active submaps include a matching submap (blue) and a growing submap (yellow). The formed 3D scan is matched and inserted into both submaps. On the second stage, if $M$ 3D scans are inserted into the matching submaps, the growing submap is switched to be the new matching submap and the old matching submap is erased. Meanwhile, a new submap (orange) is created and starts growing. During the two stages, whenever a 3D scan is formed or a new submap is created, new pose nodes are added to the SPG. 

The adjustments are done by adding an octomap \citep{hornung2013octomap} beside the original grid map. The formed 3D scan is inserted into the octomap and corresponding distance map is updated immediately. The octomap library provides an efficient method to detect changes so that the distance map can be computed efficiently. Additionally, updating octomap and distance map uses multi-threading techniques to avoid time delay caused by the distance map conversion.


\section{Experiments}\vspace{-3mm}
\label{sec:experiments}
In this section, experiments conducted in simulation and real-world are discussed. For simulation, we focus on the rotating 2D LiDAR payload which is believed to be the most challenging case compared with other types of configurations. For the real-world tests, different configurations including stationary 3D LiDAR, rotating 2D/3D LiDAR are used. \vspace{-5mm}
\subsection{Simulated Tunnel Test}
\begin{figure}[t]
\sidecaption
\includegraphics[width=0.6\textwidth]{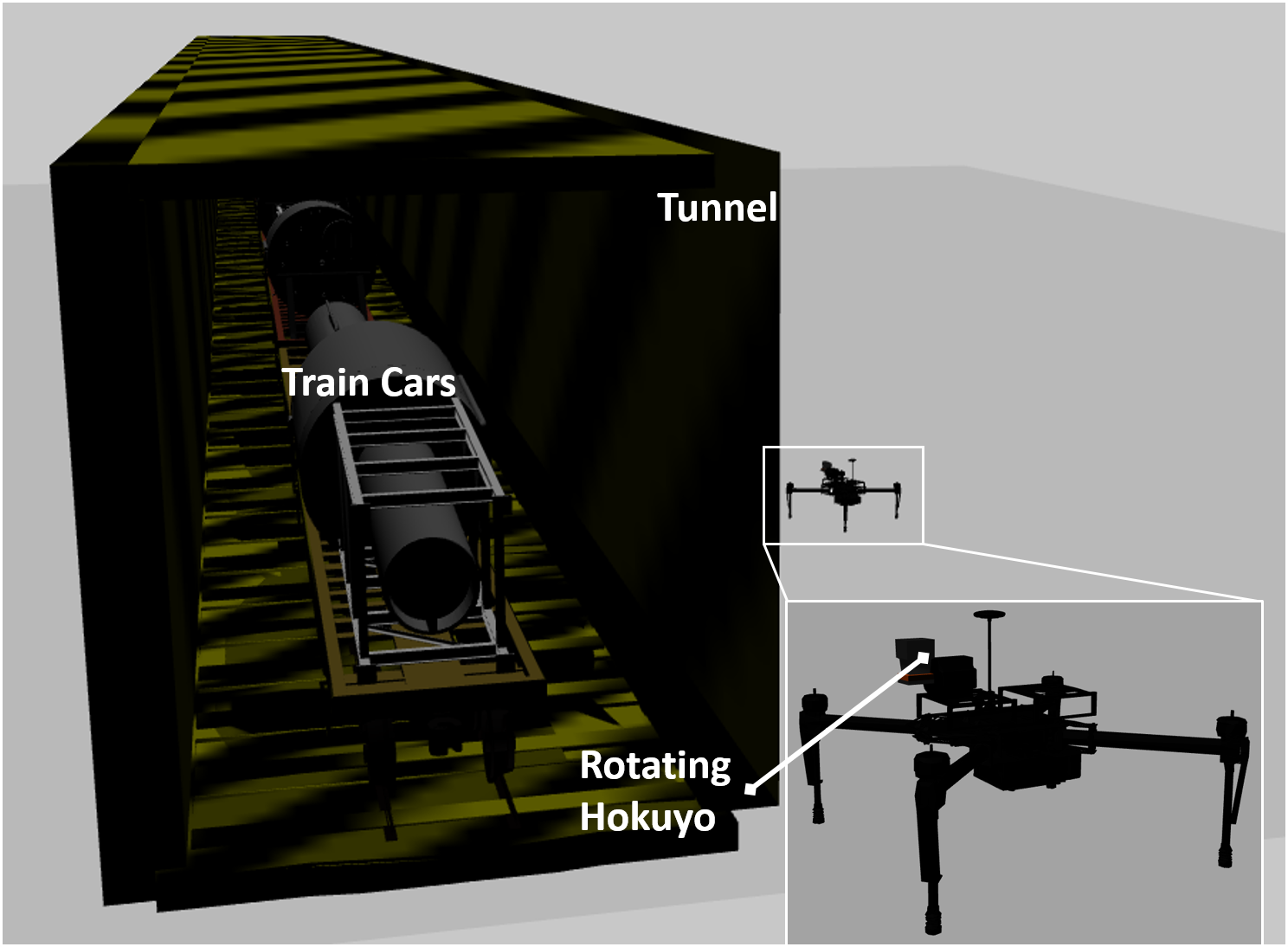}
\caption{The tunnel is of size 108m$\times$5.8m$\times$7m ($l\times w\times h$) and is built based on the DOE-PUREX nuclear tunnel wherein 8 train cars are loaded with radioactive waste. The simulated robot shares the identical sensing setup with the DJI M100. The rotating Hokuyo LiDAR is inserted from Gazebo and IMU measurements are generated by adding white noise and biases to the ground truth.}\vspace{-3mm}
\label{fig:tunnel}
\end{figure}

\vspace{-3mm}
The main task of the simulated robot is to fly through a train car tunnel (see Figure \ref{fig:tunnel}) and map its interior. The fly-through takes about 15min since we keep a relatively low velocity ($1.13$m/s at maximum) and frequently dwell so that enough laser points are collected to build submaps. Moving too fast will result in unstable localization since the submap is not well observed. This issue can be addressed using a 3D LiDAR which quickly scans many points from the environment.
\begin{figure} [t]
    \centering
    	\includegraphics[width=0.9\textwidth]{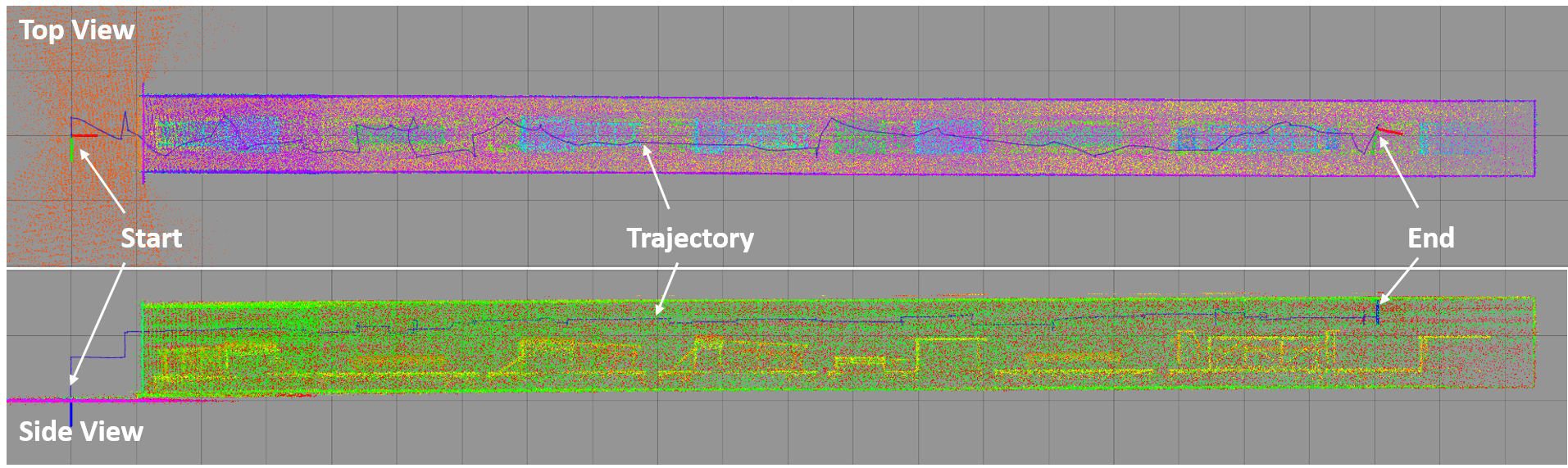}
    	\includegraphics[width=0.99\textwidth]{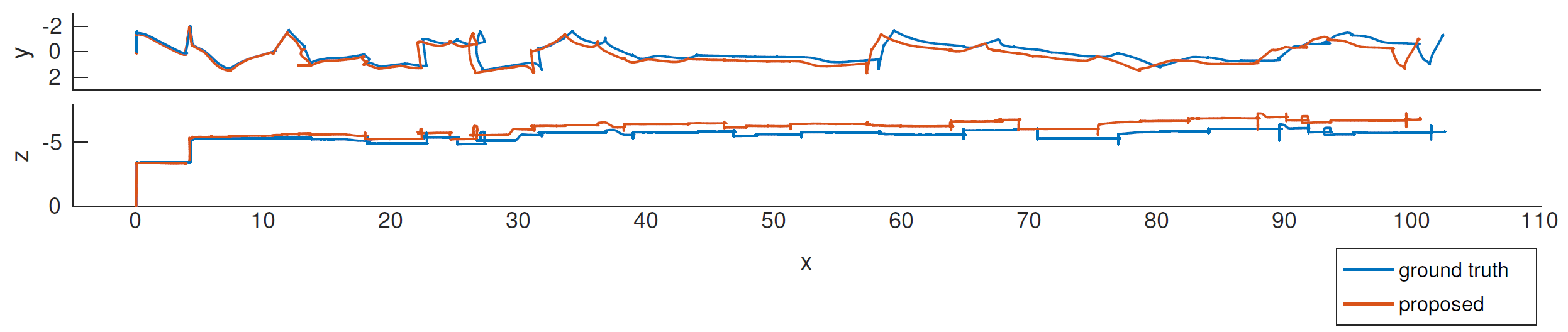}
    	\caption{\textit{Up:} Built tunnel maps along the flight through test. \textit{Down:} Comparison of the ground truth (blue) and the estimated trajectory (red). }\vspace{-3mm}
    	\label{fig:tunnel_map}
\end{figure}
\begin{figure} [t]
    \centering
    	\includegraphics[width=0.9\textwidth]{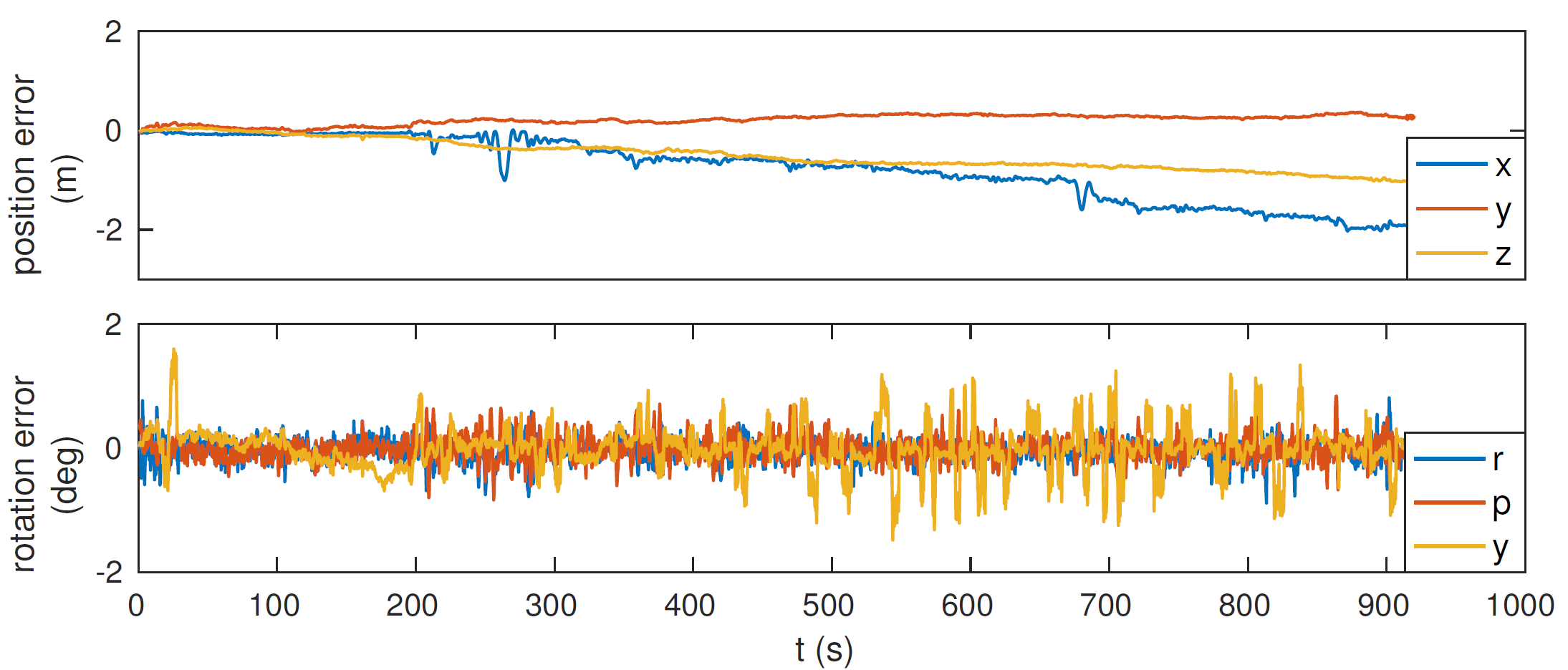}
    	\caption{A plot of pose estimation errors. The rotation error is computed by $e_{\text{rotation}} = \log\left(q_{\text{estimated}}^{-1}\cdot  q_{\text{groundtruth}}\right)$, 
where $q\in S(3)$ is a unit quaternion. The $\log(\cdot)$ function maps a unit quaternion to an angle axis vector in $so(3)$.}\vspace{-3mm}
    	\label{fig:error}
\end{figure}

The built map is visualized in Figure \ref{fig:tunnel_map} (voxel filtered with resolution 0.1m). 
In simulation, we are able to compare the estimated poses with the ground truth. From Figure \ref{fig:error}, the maximum position error in three axes is observed to be 2.0m, 0.37m and 1.11m near the end of the flight. Particularly, drift along $x$-axis is the largest, which is because the number of points on train cars to estimate $x$ is relatively small than that on side walls or ceiling to estimate $y$ and $z$. In other words, $x$-axis is under-constrained. The total traversed distance is $165$m and the translational drift rate is $1.34\%$. The rotation estimation, differently, is more consistent and the averaged error in roll, pitch and yaw are $0.14^{\circ},\; 0.15^{\circ},\;0.24^{\circ}$. There are error peaks in yaw due to occasional low quality scan matching but can be quickly recovered. The rotational drift rate is $0.003^{\circ}$/m.
\subsection{Real World Test}\vspace{-3mm}
\begin{figure}[t]
\centering
\includegraphics[width=0.99\textwidth]{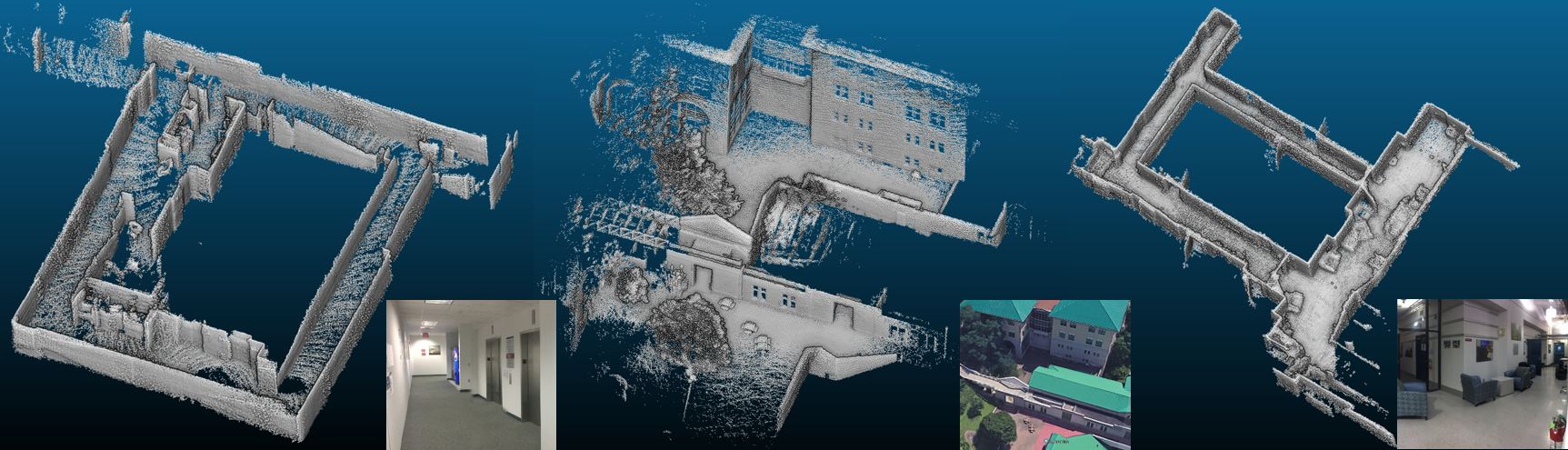}
\caption{Three tests are conducted in indoor and outdoor environments. \emph{Left}: Test with a fixed 3D LiDAR in a hallway loop. \emph{Middle}: Test with a rotating 3D LiDAR around the CMU patio. \emph{Right}: Test with a rotating 2D LiDAR in a corridor loop.}
\label{fig:results}
\end{figure}
Real-world experiments are carried out on multiple platforms: (1) fixed VLP-16 (10Hz, range 100m) with an i7 (2.5Ghz) computer, (2) rotating VLP-16, (3) rotating Hokuyo (40Hz, range 30m) with a 2.32GHz ARM processor. 

The first experiment is conducted inside a corridor loop (see Figure \ref{fig:results} left). In this test, the VLP-16 is put horizontally and the rotating speed is set to be zero so that the LiDAR is stationary. We found that although the VLP-16 measures 3D structures, its $20^{\circ}$ FOV is still not enough to reliably estimate height. The main reason is that inside the narrow corridor, most laser points come from side walls instead of the ground and ceiling. As a result, a larger drift in height is observed when the robot revisits the origin and more time is needed to detect the loop-closure. 

The second test is carried out around the patio on the CMU campus. Again the VLP-16 is used and the motor rotating speed is set to be 30 rpm. Since this is a larger area, the distance map used for localization has a coarser resolution of 0.3m and is constrained within a 40m$\times$40m$\times$40m bounding box. 

In the last test, the robot maps a narrow hallway with 1.1m width at minimum. To ensure enough laser points are collected, the robot is manually carried and moved slowly ($\approx 0.5$m/s). This time only small drifts in height is observed before closing the loop. This is because by rotating the LiDAR, the robot obtains wider FOV, which could significantly improve the mapping performance.

It is important to point out that only a few parameters (listed in Table \ref{tab:param}) are changed in the above 3 cases. For localization related parameters, $\sigma_a$ and $\sigma_g$ characterize the noise level of IMU. Distance map resolution are chosen according to the scale of environment. The maximum range of distance map sets a limit on the distance the robot will see. For mapping related parameters, $M$ and $N$ are as described in Figure \ref{fig:graph} and $I$ governs how often the background SPG got optimized. 
\vspace{-3mm} 
\begin{table} [t]
\label{tab:param}
\caption{Parameters that need to be changed when switching platforms.}
\begin{tabular}{p{5.5cm}p{5.5cm}}
\hline\noalign{\smallskip}
Localization related Params & Mapping related Params \\
\noalign{\smallskip}\svhline\noalign{\smallskip}
accelerator noise $\sigma_a$ & \# of scans per accumulation $N$\\
gyroscope noise $\sigma_g$ & \# of scans per submap $M$ \\
distance map resolution & \# of scans per optimization $I$\\
max range of distance map \\
\noalign{\smallskip}\hline\noalign{\smallskip}
\end{tabular}\vspace{-5mm}
\end{table}
\section{Conclusions}\vspace{-3mm}
\label{sec:conclusion}
In this paper, the proposed algorithm is shown to allow different LiDAR configurations to be handled in a unified framework with only a few parameters need to be tuned, which simplifies the development and application process. Some key insights obtained from the experiments are:
\begin{itemize}  
  \item The FOV of a LiDAR matters. A fixed 3D LiDAR is simple to set up but has quite a limited vertical FOV, which results in unreliable height estimation. In our experiments, the LiDAR has to be pitched down to capture more ground points. A rotating LiDAR has significantly wider FOV and is observed to be more robust to different environments. However, the rotating motor has to be carefully designed to ensure continuous data streaming and accurate synchronization. For example, an expensive slip-ring mechanism is needed to achieve continuous rotation. 
  \item Moving speed is critical in the case of rotating 2D LiDAR. In our tests, a low speed is necessary so that the laser scanner can accumulate enough points to update a submap. Moving too fast may lead to unstable localization. In the case of 3D LiDAR, a low moving speed is not a crucial requirement. 
  \item The choice of submap resolution will affect memory usage, computational complexity and mapping accuracy. From our experience, low resolution submap has low memory usage and is faster to query data from. However, that will sacrifice the map accuracy. On the other hand, higher resolution consumes more memory but doesn't necessarily improve the map accuracy. Therefore, the resolution has to be chosen carefully through trial and error.
\end{itemize}\vspace{-5mm}

\section{Acknowledge}\vspace{-3mm}
The authors are grateful to Sam Zeng, Yunfeng Ai and Xiangrui Tian for helping with the UAV development and the mapping experiments, to Matthew Hanczor and Alexander Baikovitz for building the DOE-PUREX tunnel model. This work is supported by the Department of Energy under award number DE-EM0004478. \vspace{-3mm}
\bibliographystyle{styles/spbasic}
\bibliography{mybib}

\end{document}